\documentclass[runningheads]{llncs}
\usepackage{amsmath,amssymb}
\usepackage{graphicx}
\usepackage{subcaption}
\usepackage{booktabs}
\usepackage{xcolor}
\begin{document}
\title{Wise Sliding Window Segmentation: A classification-aided approach for trajectory segmentation}
\titlerunning{Wise Sliding Window Segmentation}
%
\author{Mohammad Etemad\inst{1}
\and Zahra Etemad\inst{3}\and Am\'ilcar Soares\inst{1} \and Vania Bogorny\inst{4} \and Stan Matwin\inst{1}\inst{2}
\and Luis Torgo\inst{1}}
 
\institute{
Institute for Big Data Analytics, Dalhousie University, Halifax,
\and
Institute for Computer Science, Polish Academy of Sciences, Warsaw \and 
Bu-Ali Sina University, Hamedan, Iran
\and PPGCC - Universidade Federal de Santa Catarina (UFSC), Brazil}

\authorrunning{M. Etemad et al.}

\maketitle              
\begin{abstract}
Large amounts of mobility data are being generated from many different sources, and several data mining methods have been proposed for this data. One of the most critical steps for trajectory data mining is segmentation. 
This task can be seen as a pre-processing step in which a trajectory is divided into several meaningful consecutive sub-sequences. This process is necessary because trajectory patterns may not hold in the entire trajectory but on trajectory parts.
In this work we propose a supervised trajectory segmentation algorithm, called Wise Sliding Window Segmentation (WS-II). 
It processes the trajectory coordinates to find behavioral changes in space and time, generating an error signal that is further used to train a binary classifier for segmenting trajectory data. 
This algorithm is flexible and can be used in different domains. We evaluate our method over three real datasets from different domains  (meteorology, fishing, and individuals movements), and compare it  with four other trajectory segmentation algorithms: OWS, GRASP-UTS, CB-SMoT, and SPD. 
We observed that the proposed algorithm achieves the highest performance for all datasets  with statistically significant differences in terms of the harmonic mean of purity and coverage.

\keywords{Trajectory Segmentation \and Spatio-temporal Segmentation\and Trajectory Partition\and Supervised Trajectory Segmentation}

\end{abstract}

\section{Introduction}

An essential task for mobility data mining is trajectory segmentation. Different to classical data mining,  in trajectory data mining, the attributes/features are extracted from subtrajectory parts. The partitioning is necessary because a mobility pattern, in general, does not hold for the entire trajectory, but for subtrajectory parts. Therefore, the segmentation process becomes one of the most critical pre-processing steps for trajectory data mining.

Trajectory segmentation is the process of splitting a given trajectory into several homogeneous segments regarding some criteria. This task plays a pivotal role in trajectory mining since it affects the features of each segment, as the features may depend on the size of the trajectory segment, independently of the application domain, such as fishing detection \cite{soares2015grasp}, animal behavior \cite{soares2018semi,analytic}, tourism \cite{feng2017poi2vec}, traffic dynamics \cite{etemad2018predicting,analytic,varlamis2019network,soares2019crisis}, vessel movement patterns \cite{carliniuncovering} etc.

A trajectory is a sequence of points located in space and time, and different criteria can be used to split trajectories.
There are several approaches that can be used for trajectory segmentation such as CB-SMoT \cite{Palma2008}, 
SPD \cite{zheng2011StayPointDetection}, WK-Means \cite{Leiva2013}, GRASP-UTS \cite{soares2015grasp}, TRACLUS \cite{lee2007trajectory}, OWS \cite{etemad2019trajectory}, etc. 
Different to previous approaches where no training step is performed, we propose in this paper a supervised strategy to segment trajectory data. To the best of our knowledge this is the first approach that actually learns partitioning positions (i.e., the last trajectory point of a segment) from trajectory data characteristics for a given application domain. 
The main advantage of this supervised strategy is that the transitioning characteristics of a behavior change can be learned from the training data. 
The model built to forecast partitioning positions is further used to segment trajectories. After that, a majority vote strategy decides the proper location to place a partitioning position.

In summary, the main contributions of this work include: (i) a method for producing training data from partitioning positions on a labeled trajectory;  (ii) a method to decide when a partitioning position occurs in a trajectory; and (iii) an empirical study comparing WS-II and several baselines for segmentation.

This paper is organized as follows. Section \ref{Definitions} shows the definitions necessary to describe our trajectory segmentation method. 
In Section \ref{relatedwork}, the related works are described. In Section \ref{proposedmethod}, we propose WS-II with details. In Section \ref{experiments}, we applied the proposed method and other trajectory segmentation algorithms on three datasets and reported their performance results. Finally, we conclude our work in Section \ref{Conclusion}.

\section{Definitions} 
\label{Definitions}

In this section we present the basic concepts related to trajectories and used throughout this paper.

The trajectory of a moving object $o$ can be described by a time ordered sequence of locations the object has visited. We call these locations, \textit{trajectory points}.

\subsubsection*{\textbf{Trajectory Point}}
A \textit{trajectory point}, $l^o_i$, is the location of object $o$ at time $i$, and is defined as,
\begin{equation}
l^o_i=\langle x^o_i, y^o_i \rangle
\label{not:1}
\end{equation}
\noindent where $x^o_i$  is the longitude of the location which varies from 0$^{\circ}$ to $\pm 180^{\circ}$, while $y^o_i$ is the latitude which varies from 0$^{\circ}$ to $\pm 90^{\circ}$.

\subsubsection*{\textbf{Raw Trajectory}}

A \emph{raw trajectory}, or simply \emph{trajectory}, is a time-ordered sequence of trajectory points of some moving object $o$,

\begin{equation}
\tau^o=\langle l^o_0,l^o_{1},..,l^o_n\rangle
\label{not:2}
\end{equation}

\subsubsection*{\textbf{Segment or Subtrajectory}}
is a set of consecutive trajectory points belonging to a raw trajectory $\tau^o=\langle l^o_0,l^o_{1},..,l^o_n\rangle$,

\begin{equation}
 s^o=\langle l^o_j,\cdots ,l^o_{k}\rangle,\ \ j \geq 0,\ k\leq n \  \mathrm{and} \ \ s^o\subset \tau^o
\label{not:2_1_2}
\end{equation}

The process of generating segments from a trajectory is called \emph{Trajectory Segmentation ($TS$)}. The most common way of defining TS involves splitting a raw trajectory into a set of non-overlapping segments. More formally:

\subsubsection*{\textbf{Trajectory Segmentation}}
Given a raw trajectory $\tau^o=\langle l^o_0,l^o_{1},..,l^o_n\rangle$, we define a sequence of segments $S = \langle s^o_0, \cdots , s^o_k\rangle$, such that
\begin{equation}
 \forall_{s^o_i,s^o_{i+1}\in S} \ s^o_i=\langle l^o_p,\cdots,l^o_{p+t}\rangle,\  s^o_{i+1}= \langle l^o_{p+t+1},\cdots,l^o_{p+t+u}\rangle
\label{not:2_1_3}
\end{equation}

and 

\begin{equation}
s^o_0=\langle l^o_0,\cdots,l^o_i\rangle,\  s^o_{k}= \langle l^o_{j},\cdots,l^o_{n}\rangle
\label{not:2_1_4}
\end{equation}

Equation \ref{not:ts} shows the input and output of the trajectory segmentation process, where $\tau$ is a raw trajectory which contains $n$ trajectory points, and $S$ is the set of all segments generated from $\tau$ using $TS$.

\begin{equation}
TS: \tau \longrightarrow S  \ \      ,|\tau|=n+1,\ \  |S|=k+1
\label{not:ts}
\end{equation}
In this notation,  $n+1$ is the number of trajectory points and $k+1$ is the number of segments resulting from applying $TS$ to the trajectory.

We call a trajectory point at the end of each segment as \textit{partitioning position}. This means that the result of applying $TS$ to a trajectory, $S$ contains $k$ partitioning positions.

\subsubsection*{\textbf{Problem Definition}}
Given a raw trajectory $\tau^o$, we would like to generate a sequence of segments $S = \langle s^o_0, \cdots , s^o_k\rangle$ so that each $s^o_i$ satisfies a certain homogeneity criteria for a given application domain.
To evaluate the performance of the generated $S$, we rely on the knowledge of an expert user to provide a set of semantic tuples $sl_i=(sid,label)$ where $sid$ identifies a segment $s_i$ of a trajectory, generated by the expert user, and $label$ is a semantic label attached by the expert to this segment, such as for instance, a transportation mode or status of fishing or non-fishing.

\begin{figure*}[htb]
\centerline{\includegraphics[width = 0.8\textwidth]{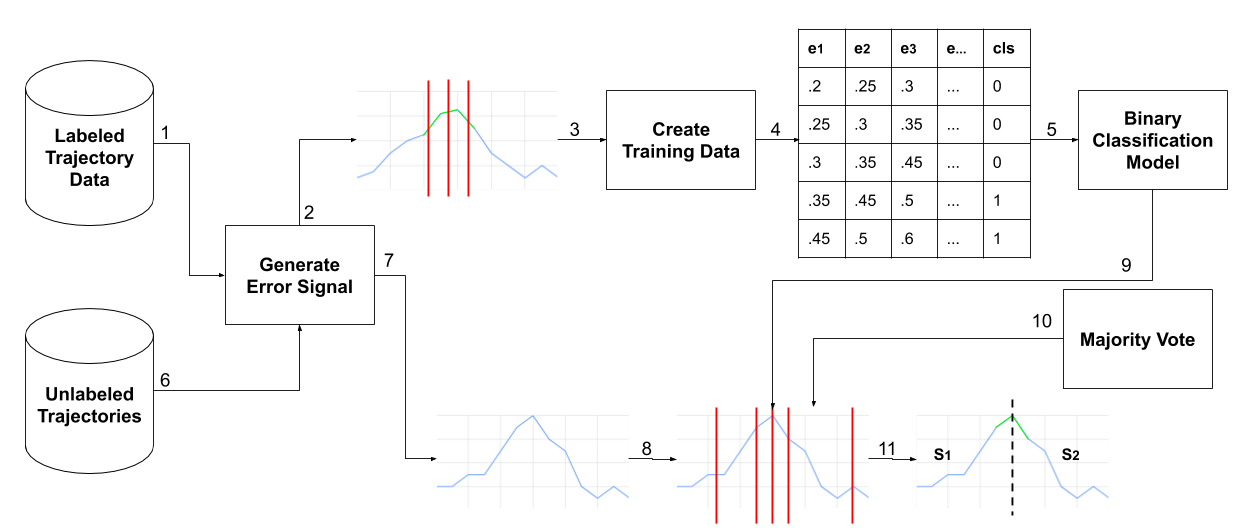}}
\caption{A high level view of wise sliding window segmentation algorithm.}
\label{fig:wsii}
\end{figure*}

\section{Related works}
\label{relatedwork}

In this section, we give an overview of several methods for trajectory segmentation.
Warped K-Means (WK-Means), which is a general-purpose segmentation algorithm based on K-Means \cite{macqueen1967some}, is introduced in \cite{Leiva2013}. 
It modifies the K-Means algorithm by minimizing a quadratic error (cost function) while imposing a sequential constraint in the segmentation step. 
Since WK-Means imposes a hard sequential constraint, segments can be updated while new samples arrive without affecting too much the previous clustering configuration \cite{Leiva2013}. 
This algorithm receives 
the number of segments to be found on the data ($k$). Having such input parameter is the main limitation of using it in domains where the number of segments is not pre-defined or is dynamic.

The Stay Point Detection (SPD) \cite{zheng2011StayPointDetection} is a simple algorithm that follows the idea that between each two-movements, there is a stop. 
SPD applies a distance threshold ($\theta_d$) and a time threshold $\theta_t$ so that a moving object which spends more than $\theta_t$ time in the neighborhood of $\theta_d$ belongs to a stay point. Hence, each stay point identifies a segment, and the trajectory points between two stay points are generated in another segment. 

An extension of DB-SCAN \cite{ester1996density}, CB-SMoT detects stops and moves segments in a trajectory\cite{Palma2008}.
The original definitions of a $\epsilon-$neighborhood and minimum points in DB-SCAN are altered so that CB-SMoT utilizes spatial and temporal aspects of trajectories. 
CB-SMoT works based on the trajectory speed, and the stop points are consecutive trajectory points where the moving object has a lower speed.

TRACLUS \cite{lee2007trajectory} detects dense regions with the same line segment characteristics. This clustering algorithm has two steps: (i) partitioning of the trajectory to line segments; and (ii) clustering these lines. 
A cost function based on the Minimum Description Length (MDL) principle is applied in the first step to split a trajectory into its line segments. It considers three trajectory segment's attributes:(i) parallel distance, (ii) perpendicular distance, and (iii) angular distance.
Clustering line segments using DB-SCAN is run in the next step \cite{lee2007trajectory}.

GRASP-UTS is an unsupervised trajectory segmentation algorithm that benefits from the Minimum Description Length (MDL) principle to build the most homogeneous segments. First, GRASP-UTS generates random landmarks. Then, it builds homogeneous segments by swapping the trajectory points across temporally-ordered segments and adjusting the landmarks based on its cost function's value \cite{soares2015grasp}.
GRASP-UTS can apply additional features on top of the raw trajectories to perform trajectory segmentation.

 The OWS (Octal Window Segmentation) algorithm is based on computing the error signal generated by measuring the deviation of a middle point of an octal window \cite{etemad2019trajectory}. 
The intuition behind OWS is that when a moving object changes its behavior, this shift may be detected using only its geolocation over time \cite{etemad2019trajectory}. 
OWS uses interpolation methods to find the estimated position of the moving object, i.e., where it is supposed to be if its behavior does not change. 
Then, OWS compares the real position of the moving object with the estimated one, creating an error signal. 
With such a procedure, it is possible to determine where the moving object changed its behavior and to use this information to create segments.

In this work, we extend the idea of OWS by using a configurable sliding window for interpolating points and a supervised strategy for deciding where partitioning positions should be placed. 
Unlike all previous segmentation algorithms, WS-II is supervised. 
This means that WS-II is able to learn the variations in the error signal generated by interpolation techniques which characterize partitioning positions over consecutive segments, avoiding in this way the decision of choosing an error threshold value (i.e., an epsilon value in the OWS) that relies on the characteristics of the domain where trajectories were collected.

\section{The proposed method}
\label{proposedmethod}

Figure \ref{fig:wsii} shows an overview of the Wise Sliding Window Segmentation (WS-II) method, which has four core procedures: Generate Error Signal, Create Training Data, Binary Classification Model, and Majority Vote. 
First, the WS-II creates the error signal from the labeled dataset, which is detailed in Section \ref{ss:errorsignal}. 
The second step is to generate the training data using the error signal, by sliding a window over its values and adding the presence or absence of a partitioning position. This part is detailed in Section \ref{ss:trainingdata}.
The third step is to train a binary classifier to recognize the partitioning positions over the sequence of error signals. This part is detailed in Section \ref{ss:binaryclassification}. Finally, unlabeled trajectories can then be segmented based on the model learned in the previous step and using the majority vote, as detailed in Section \ref{ss:majorityvote}.

\subsection{Generating the error signal}
\label{ss:errorsignal}

\begin{figure}[t]
        \centering
        \begin{subfigure}[t]{0.47\textwidth}
            \centering
            \includegraphics[width = 0.9\textwidth]{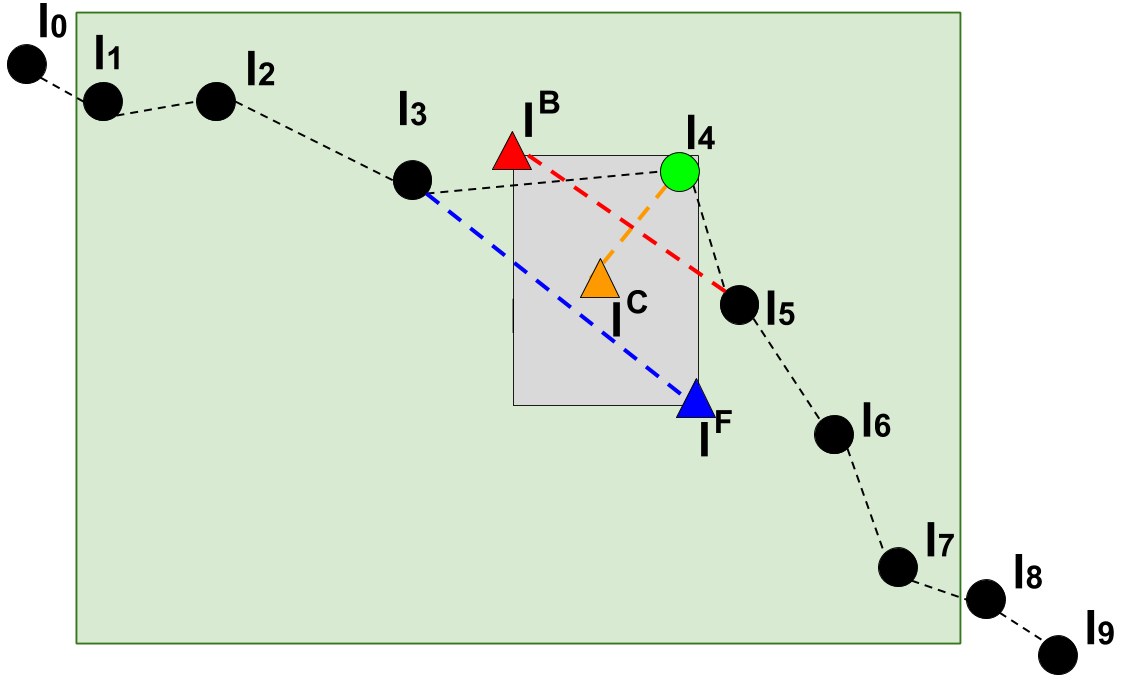}
\caption{Fig. 2a Example of error calculation where seven trajectory points (e.g., $l_1$ to $l_7$) are selected as the \textit{current sliding window} (e.g., green box). }
\label{fig:SOWexample}
        \end{subfigure}
        \hfill
        \begin{subfigure}[t]{0.45\textwidth}  
            \centering 
              \includegraphics[width = 0.9\textwidth]{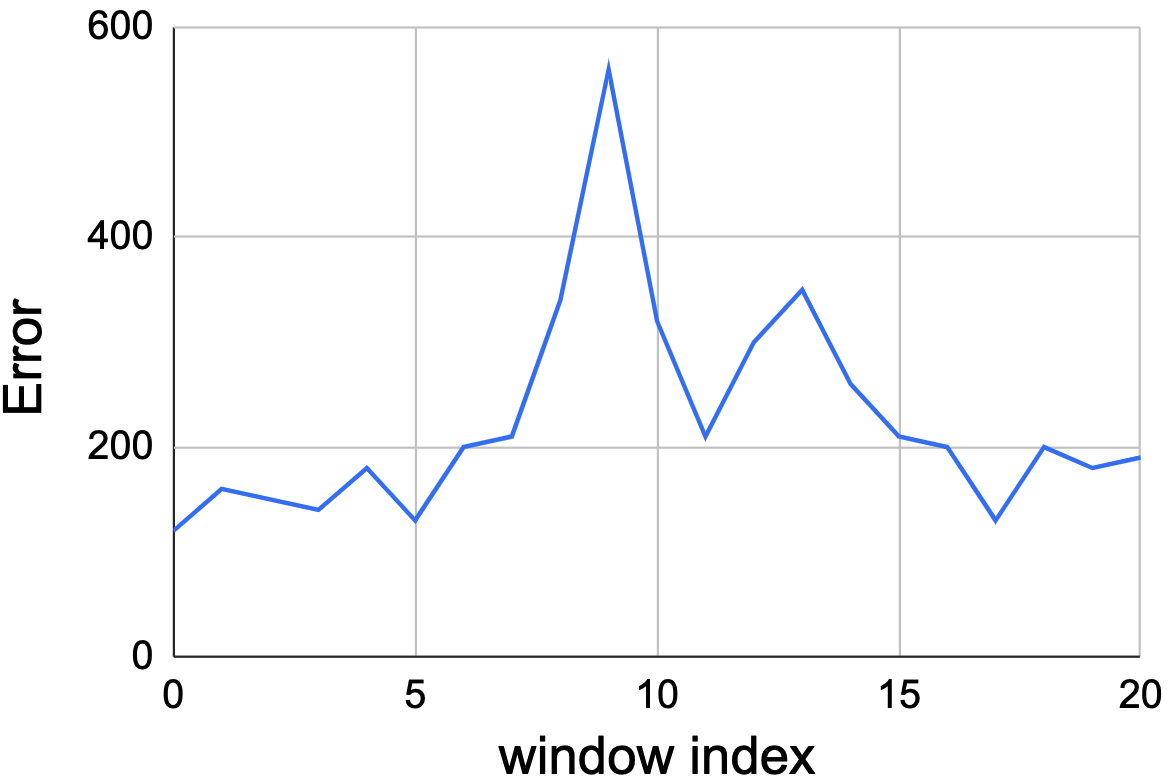}
\caption{Fig. 2b Example of error signal generated in meters for trajectory data.}
\label{fig:errorSignal}
        \end{subfigure}
        \vskip\baselineskip
\end{figure}

The first step of our proposal is similar to the OWS algorithm \cite{etemad2019trajectory}, which creates a sliding window over a trajectory to compute a signal error between trajectory points. 
For each sliding window, the error is generated by calculating the deviation of the interpolated midpoint of the window from the actual midpoint. 
This process is repeated by sliding the window by one point forward, so receiving a new trajectory point, it adds the newer point to the window set and removes the oldest point from the set. 
An example of this process is shown in Figure \ref{fig:SOWexample}.

In Figure \ref{fig:SOWexample}, the green rectangle is a sliding window of size 7, the $l^B$ (red triangle), and $l^F$ (blue triangle) are the interpolated positions. $l^F$ is generated using extrapolation on the first three points ($l_1,l_2,l_3$) and $l^B$ is generated using the last three points inside the window ($l_5,l_6,l_7$). 
The green dot ($l_4$) is assumed to be the missing point in the sliding window, while the $l^C$ (orange triangle) is generated as a middle point between $l^B$ and $l^F$.    
The distance between the midpoint ($l^C$) and the missing point ($l_4$) is called the \textit{error value} of this window. 
In the example of Figure \ref{fig:SOWexample}, the haversine distance from the estimated position $l^C$ to the real position $l_i$ is visible. 
This may indicate that the moving object's behavior has changed at position $l_4$.

 An example of the error signal from a trajectory is shown in Figure \ref{fig:errorSignal}. 
A raw trajectory with 26 points that forms 20 sliding window of size 7 (generating error for point index 4 to 23) is displayed in this example. 
The first three and last three error values are dropped.  
Window index is the index of the middle trajectory point in each window.
Figure \ref{fig:errorSignal} illustrates a situation in which they are several trajectory points (e.g., around trajectory points 8 and 12)  along the raw trajectory where the estimated positions were far from the real trajectory positions.
These boundaries are considered as potential partitioning positions for creating trajectory segments.

\subsection{Creating Training Data}
\label{ss:trainingdata}
The second core procedure of WS-II is to create a training dataset using the sequential error values extracted in the previous step. 
First, we create an array of size $q$ of error signals that will belong to the first training sample, and we use the ground truth information (i.e., if in this particular region there was a change in the behavior) to annotate the label of this sample. 
If this window includes a partitioning position, it is labeled as $1$  and $0$ otherwise. 
By receiving every new trajectory point, we remove one point from the start of our window and add the new point to the end of the window. 
Then we create our next sample by applying the same step of labeling $1$ when a partitioning position is present in the sliding window, and $0$ if it is not. 
This procedure is repeated until all the error signals are evaluated.

\begin{figure*}[bt]
        \centering
        \begin{subfigure}[b]{0.43\textwidth}
            \centering
            \includegraphics[width = 0.85\textwidth]{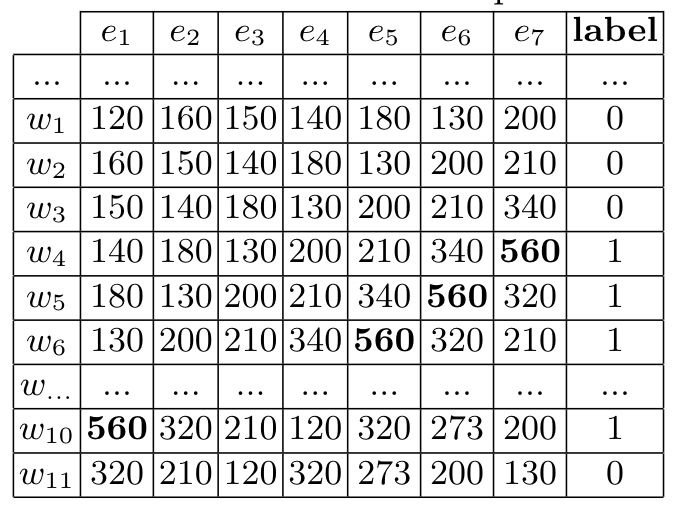}
\caption{Example of a training set generated by WS-II.}
\label{fig:training_data_example}
        \end{subfigure}
        \hfill
        \begin{subfigure}[b]{0.43\textwidth}  
            \centering 
              \includegraphics[width = 0.85\textwidth]{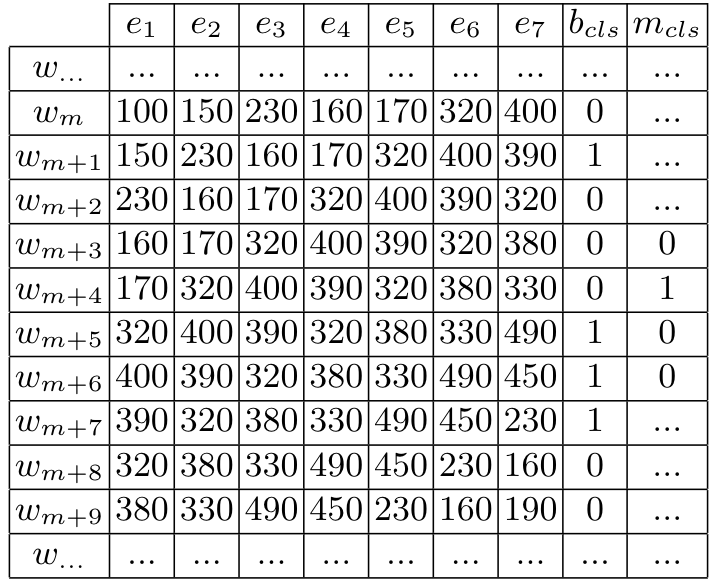}
\caption{Example of the majority vote mechanism.}
\label{fig:majorityvote}
        \end{subfigure}
        \vskip\baselineskip
\caption{Examples of data tables generated by our algorithm.}
\end{figure*}

 To understand how the labeling process works, we show an example in Figure \ref{fig:training_data_example}. In this example, the training data are created for the sliding window built with seven ($e_1$ to $e_7$) trajectory points over eleven slides (i.e., $w_1$ to $w_{11}$). 
 As can be seen in Figure \ref{fig:training_data_example}, from $w_1$ to $w_3$, there was no big change in the error signal (ranging from 120 to 340 meters). 
 In $w_4$, the value of 560 characterizes a high jump in the estimated error and actually reflects a real change in the behavior of the moving object, resulting in a positive example (i.e., there is a partitioning position) in the training data. 
 Examples from $w_4$ to $w_{10}$ are labeled as positive due to the presence of partitioning position in the sliding window.
 From $w_{11}$, the samples are again labeled as negative examples due to the absence of a partitioning position in the data.

\subsection{Binary Classification Model}
\label{ss:binaryclassification}

A binary classifier is used by WS-II to categorize each error signal sample into either a partitioning position or not.
The labeled trajectory data created in the previous step is used to generate training samples for this binary classifier so that it can classify signal samples into a class where a sliding window has a partitioning position (e.g., value 1) or a class when it does not have a partitioning position (e.g., value 0).

It was observed that the error signal has its minimum fluctuations far from a partitioning position, and it has its maximum fluctuations while transitioning from one segment to a new one.
Therefore, detecting the area that includes partitioning positions is an indicator that the behavior has changed.
We apply the binary classifier to identify these areas over a trajectory that has the highest likelihood of containing partitioning positions. 
In this work, we used a Random Forest classifier \cite{Breiman2001} 
to benefit from its bagging power while processing long window sizes faster by limiting the number of features. 
However, we emphasize that any classification model can be used in this step. 
After forecasting these transitioning areas, we use a majority vote mechanism to decide precisely where to place a partitioning position, explained in the next section.

\subsection{Majority Vote}
\label{ss:majorityvote}

At this step, we use the same sliding window of size $q$ to decide if a partitioning position occurred.
Since we are using a window slide point by point, each trajectory point can be part of $q$ sliding windows, and we classify each window using the binary classifier. 
This means that we have $q$ outputs that the binary classifier generates for a trajectory point belongs to the $q$ windows. 
Using a majority vote mechanism for these $q$ outputs leads us to the final decision: the trajectory point is a partitioning position if more than 50\% of the sampled signals are labeled as a partitioning position.

Leveraging this feature and applying the voting technique, we can have a more robust evaluation to support if a point is a partitioning position or not. The decision to identify a trajectory point as a partitioning position is supported by $q$ results, each of which contributes $1/q$ to the final decision. 
This means a misclassification of the binary classifier weights $1/q$. Although increasing $q$ can make the algorithm more robust to noise, it will make it fail to identify segments with a length smaller than $q$.
Furthermore, the algorithm is more robust against noisy points, which may happen in trajectory data due to device collection errors.

An example of the advantages of the majority vote mechanism are exemplified in Figure \ref{fig:majorityvote}, where a window with $q = 7$ was used. 
In Figure \ref{fig:majorityvote}, the column $b_{cls}$ was forecast by the binary classifier for $w_m$ to $w_{m+9}$. 
It is possible to see in Figure \ref{fig:majorityvote} that $w_{m+3}$ is decided by evaluating the $b_{cls}$ column values from $w_m$ to $w_{m+6}$ (0,1,0,0,0,1,1). The decision regarding a majority vote for $w_{m+3}$ is equal to 0 since $|\#0|=4$ and $|\#1|=3$. 
For deciding the final value of $w_{m+4}$ the lines from $w_{m+1}$ to $w_{m+7}$ are used. 
The evaluation of the set (1,0,0,0,1,1,1) through a majority vote ($|\#0|=3$ and $|\#1|=4$) results in the decision of 1 (i.e., a partitioning position occurred). 
As previously stated, such strategy makes WS-II robust against spatial jumps due to GPS error in the data collection process.

\section{Experimental Evaluation}
\label{experiments}
In this section, we evaluate the proposed method and compare it to state-of-the-art approaches. In Section \ref{Datasets}, we describe 
the datasets. In Section \ref{ExperimentSetup}, we detail the  experimental setup and  we report 
the results in Section \ref{fishing_cmp}.

\subsection{Datasets}
\label{Datasets}

We evaluate our method on three datasets. The first is a fishing dataset containing 5190 trajectory points and 153 segments, where fishing activity labels (e.g., fishing or not-fishing) were provided by specialists and used to create trajectory segments. 
The second is the Atlantic hurricane dataset, which contains 1990 trajectory points and 182 segments. The Saffir-Simpson scale was used to determine the type of hurricane, and the transitions from one hurricane-level to another was used for creating trajectory segments. 
Finally, a subset of the Geolife dataset containing 12,955 trajectory points and 181 segments was used as a third dataset. For this dataset, we use the transportation mode as the ground truth for creating the segments. 
The reason that we did not use the full Geolife data set was that some of the segmentation algorithms, such as GRASP-UTS were not able to provide segments in a reasonable time. 
We create a sample Automatic Identification System (AIS) data to debug our algorithm and test our code and made it available to public \footnote{https://github.com/metemaad/WS-II}.

\subsection{Experimental Setup}
\label{ExperimentSetup}
In this work we measure the trajectory segmentation performance using Harmonic mean of Purity and Coverage, introduced in \cite{etemad2019trajectory}. 
The use of purity and coverage for trajectory segmentation performance measurement originally is introduced in \cite{soares2015grasp}. 
We do not use clustering measures such as completeness and homogeneity since the segmentation task is different from clustering. 
In trajectory segmentation, the order of the segments is essential, and adjacent segments can come from the same cluster. 
For example, an object moving to a shopping store and going back home characterizes two segments, that would be in the same "walk" cluster.

In each experiment, we divided the dataset into ten folds, one of which is applied to tuning/training the algorithm and the rest to testing its performance. 
Each fold contains different trajectories of different moving objects; therefore, we individually segment each trajectory and report the average results.

Since we divide data into ten folds, we calculate ten values for the Harmonic means. A boxplot is used to show the visual difference between these ten values for each algorithm, Figure \ref{fig:compareall}. Although the boxplot can show the difference between the performance of algorithms, we perform a Mann Whitney U test (having only ten numbers, we could not prove the data follows normal distribution, so we did not use T-test) to show that the difference between the median of each set is not generated randomly.

The state of the art methods that we compare to our approach requires some parameterization. 
The input parameter values estimation for GRASP-UTS was using a grid search with all combinations of values reported in \cite{soares2015grasp}.
For the SPD algorithm, we used the suggested parameters on the original paper for the subset of Geolife dataset, and for the rest of datasets we used a grid search to find the best parameters. 
For CB-SMoT, we applied a grid search to tune parameters using the parameter tuning fold. 
For OWS, we have tested the four kernels (e.g., random walk, kinematic, linear, and cubic) and used the same strategy reported in \cite{etemad2019trajectory} to find the best value of $epsilon$.
We decided only to report the random walk kernel findings since it obtained good results for all datasets. 
For a fair comparison between OWS and WS-II, we only report the WS-II results with the random walk kernel. 
Details regarding the input parameter values ranges for all algorithms can be found in the following link \footnote{https://github.com/metemaad/WS-II}

\subsection{Results and discussion}
\label{fishing_cmp}
\begin{figure}[h]
        \centering

            \includegraphics[width = 0.8\textwidth]{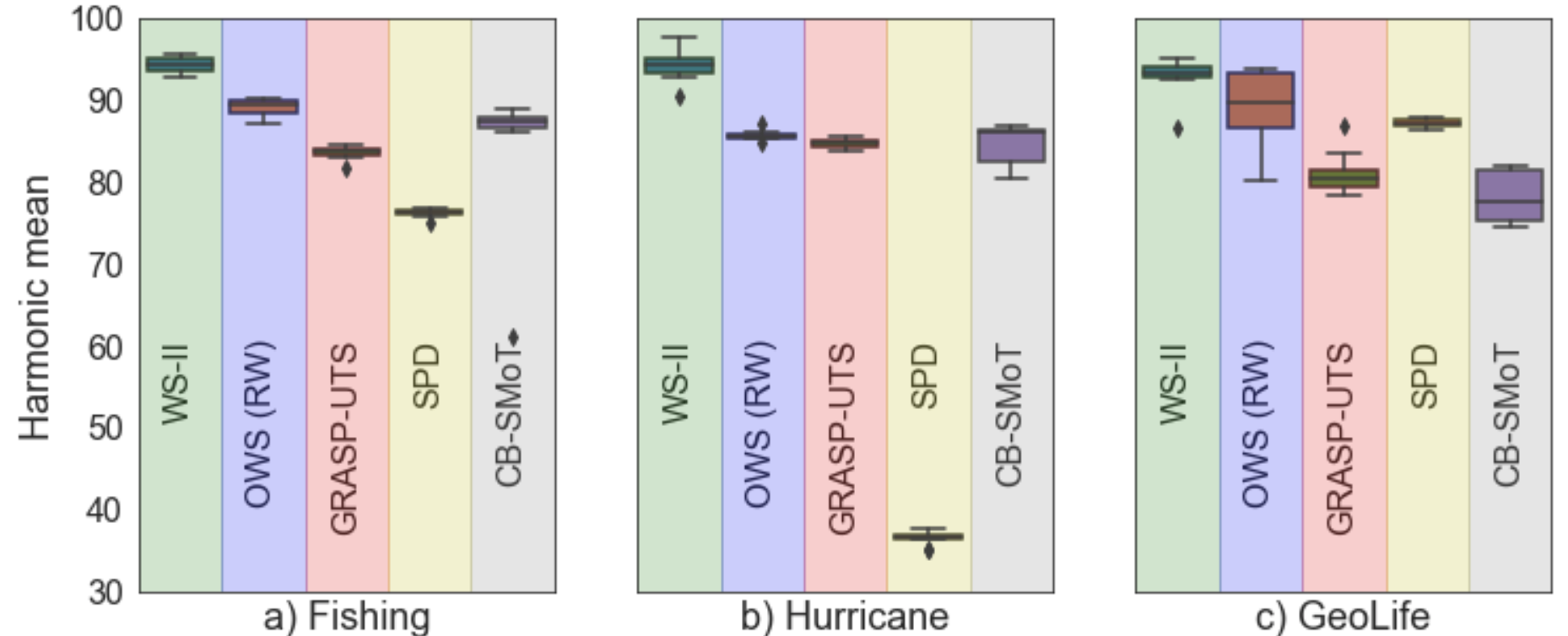}
\caption{\small Our proposed method outperforms CB-SMoT, SPD, GRASP-UTS, and OWS with Random Walk kernel on Fishing, Hurricane and Geolife dataset}
\label{fig:compareall}
\end{figure}

Figure \ref{fig:compareall}.a displays the results of executing different segmentation algorithms on the Fishing dataset. A Mann Whitney U test indicated that WS-II produces statistically significant higher median ($mean$ =94.32, $std$=0.9) harmonic mean for trajectory segmentation comparing to OWS with random walk kernel ($p_{value}$ = 9.133e-05, $mean$=89.04, $std$=1.03). Therefore, the proposed method achieved better performance in comparison to other trajectory segmentation methods. 

A fishing activity is characterized by several ship turns. We believe that WS-II had a better result when compared with the other algorithms because of its capability to analyze not only a single trajectory point, but a larger region (i.e., a larger sliding window size). By analyzing a larger region, WS-II will only place a partitioning position when a partitioning position actually occurred (e.g., learned from the training data). 
Since a single turn is not enough to characterize a fishing activity, WS-II's strategy of analyzing a larger window is more robust in learning such behavior.

In this experiment, we compare five trajectory segmentation algorithms: CB-SMoT, SPD, GRASP-UTS, OWS with Random Walk kernel, and our proposed trajectory segmentation algorithm (WS-II) on Atlantic hurricane dataset. Figure \ref{fig:compareall}.b shows that WS-II performed better than all other algorithms.
A Mann Whitney U test indicated that WS-II produces statistically significant higher median ($mean$=94.68, $std$=2.23) harmonic mean for trajectory segmentation comparing to OWS with random walk kernel ($p_{value}$=9.1e-05, $mean$=85.67, $std$=0.59).

In this experiment, we applied all the segmentation algorithms on a subset of Geolife containing ten different users. Each user's trajectory creates a fold and we use one fold to tune up our algorithm each time.
Figure \ref{fig:compareall}.c depicts our experiment results.
Moreover, 
a Mann Whitney U test supports the claim that WS-II produces statistically significant higher median ($mean$ =92.8, $std$=2.11) harmonic mean for trajectory segmentation comparing to OWS with random walk kernel ($p_{value}$ = 0.00065, $mean$=88.94, $std$=5.06). 
 
 In the Geolife dataset, there are two major types of movement: (1) fast movements of buses, trains, and  cars; and (2) slow movements of walk and bike, which have a random nature. The selection of a random walk seems to be a reasonable decision in this dataset because as long as a moving object moves slowly, the random walk kernel seems to reproduce the random nature of the movement. On the other hand, for the moving object that travels fast, the behavior of random walk is similar to a linear interpolation kernel in terms of direction because the direction variation decreases. 

\section{Conclusions}
\label{Conclusion}
In this paper we presented a supervised method for trajectory segmentation named Wise Sliding Window Segmentation (WS-II), that uses a trained model for deciding where partitioning positions should be placed. With the majority voting strategy the method becomes more robust to noise points and avoiding unnecessary partitions.
The experimental results show that WS-II achieves better performance in terms of a harmonic mean of purity and coverage when compared with state-of-art trajectory segmentation algorithms in three datasets of different domains. One limitation of WS-II, which is a limitation for all learning methods, is that several domains do not have a labeled dataset where the patterns of movement behavior change can be learned. Although there are tools in the literature that encourage and assist the user in the process of labeling trajectory data \cite{vista}, most trajectory datasets still do not provide any type of ground truth for validating supervised methods.
As future work, we would like test how this algorithm performs with different sample sizes for training, i.e.,  how large the labeled data needs to be to find good results.
 
 \subsubsection*{Acknowledgments}
This work was financed by the Brazilian Agencies CNPq, CAPEs(Project Big Data Analytics [CAPES/PRINT process number 88887.310782/2018-00]), FAPESC (Project Match co-financing of H2020 Projects - Grant 2018TR 1266); the European Union’s Horizon 2020 research and innovation programme under Grant Agreement 777695 (MASTER1); and the Natural Sciences and Engineering Research Council of Canada (NSERC).

\bibliographystyle{splncs04}
\bibliography{refs}
\end{document}